\documentclass[10pt,twocolumn,letterpaper]{article}

\usepackage{wacv}
\usepackage{times}
\usepackage{epsfig}
\usepackage{graphicx}
\usepackage{amsmath}
\usepackage{amssymb}
\usepackage{caption}
\newcommand{\mbf}[1]{\mathbf{#1}}
\DeclareMathOperator*{\argmin}{argmin}

\usepackage{booktabs}

\usepackage{color}

\newcommand{\mth}{\boldsymbol{\theta}}
\newcommand{\mbt}{\boldsymbol{\beta}}

\usepackage[pagebackref=true,breaklinks=true,letterpaper=true,colorlinks,bookmarks=false]{hyperref}

\usepackage{verlab}
\usepackage{fancyhdr}

\wacvfinalcopy 

\def\wacvPaperID{396} 

\ifwacvfinal\fi
\begin{document}


\title{Do As I Do: Transferring Human Motion and Appearance \\between Monocular Videos with Spatial and Temporal Constraints}

\author{Thiago L. Gomes$^1$ \hspace{1cm} Renato Martins$^{1,2}$ \hspace{1cm} Jo\~ao Ferreira$^1$ \hspace{1cm} Erickson R. Nascimento$^1$\\
\hspace{-1cm} $^1$Universidade Federal de Minas Gerais (UFMG), Brazil \hspace{0.5cm} $^2$INRIA, France\\
{\tt\small \{thiagoluange,renato.martins,joaoferreira,erickson\}@dcc.ufmg.br}
}


\maketitle
\thispagestyle{fancy}
\fancyhf{}
\chead{{To appear in Proceedings of the IEEE Winter Conference on Applications of Computer Vision (WACV) 2020 \\ The final publication will be available soon.}}

\begin{abstract}
	
	Creating plausible virtual actors from images of real actors remains one of the key challenges in computer vision and computer graphics. Marker-less human motion estimation and shape modeling from images in the wild bring this challenge to the fore. Although the recent advances on view synthesis and image-to-image translation, currently available formulations are limited to transfer solely style and do not take into account the character's motion and shape, which are by nature intermingled to produce plausible human forms. 
	In this paper, we propose a unifying formulation for transferring appearance and retargeting human motion from monocular videos that regards all these aspects. Our method synthesizes new videos of people in a different context where they were initially recorded. Differently from recent appearance transferring methods, our approach takes into account body shape, appearance, and motion constraints. The evaluation is performed with several experiments using publicly available real videos containing hard conditions. Our method is able to transfer both human motion and appearance outperforming state-of-the-art methods, while preserving specific features of the motion that must be maintained (\eg, feet touching the floor, hands touching a particular object) and holding the best visual quality and appearance metrics such as Structural Similarity (SSIM) and Learned Perceptual Image Patch Similarity (LPIPS). 		

\end{abstract}

\section{Introduction}

Humans start learning early in their lives to recognize human forms and make sense of what emotions and meaning are being communicated by human movement. We are, by nature, specialists in the human form and movement analysis. Even for a meticulous artist, it may be hard to capture in a purely manual approach the fine details of human form and motion. 
 Human form and motion estimation is at the core of a wide range of applications including entertainment, graphic animation, virtual and augmented reality, to name a few.

\begin{figure}[!t]
	\includegraphics[width=1\linewidth]{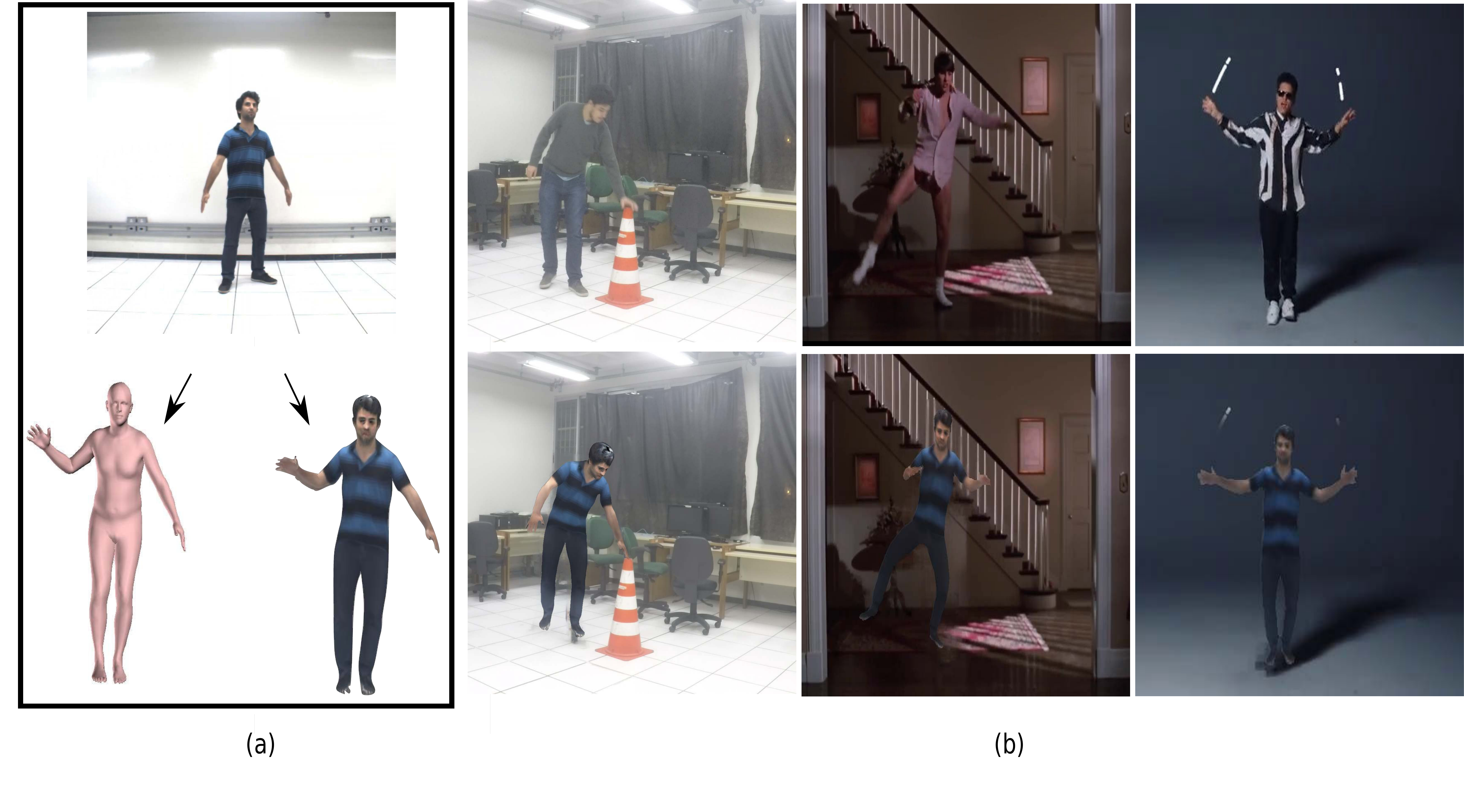}
	\captionof{figure}{\small Overview of the motion and appearance transfer from a target video to different videos. After reconstructing a model for the target human (shown in (a)), we transfer his shape and motion to different videos as shown in (b). {\bf Top row}: video with the source motion. {\bf Bottom row}: New video with the retargeted motion and appearance of the target human model.}
\label{fig:teaser}  
\vspace{-0.2cm}
\end{figure}

\begin{figure}[t!]
	\centering
	\includegraphics[width=0.81\linewidth]{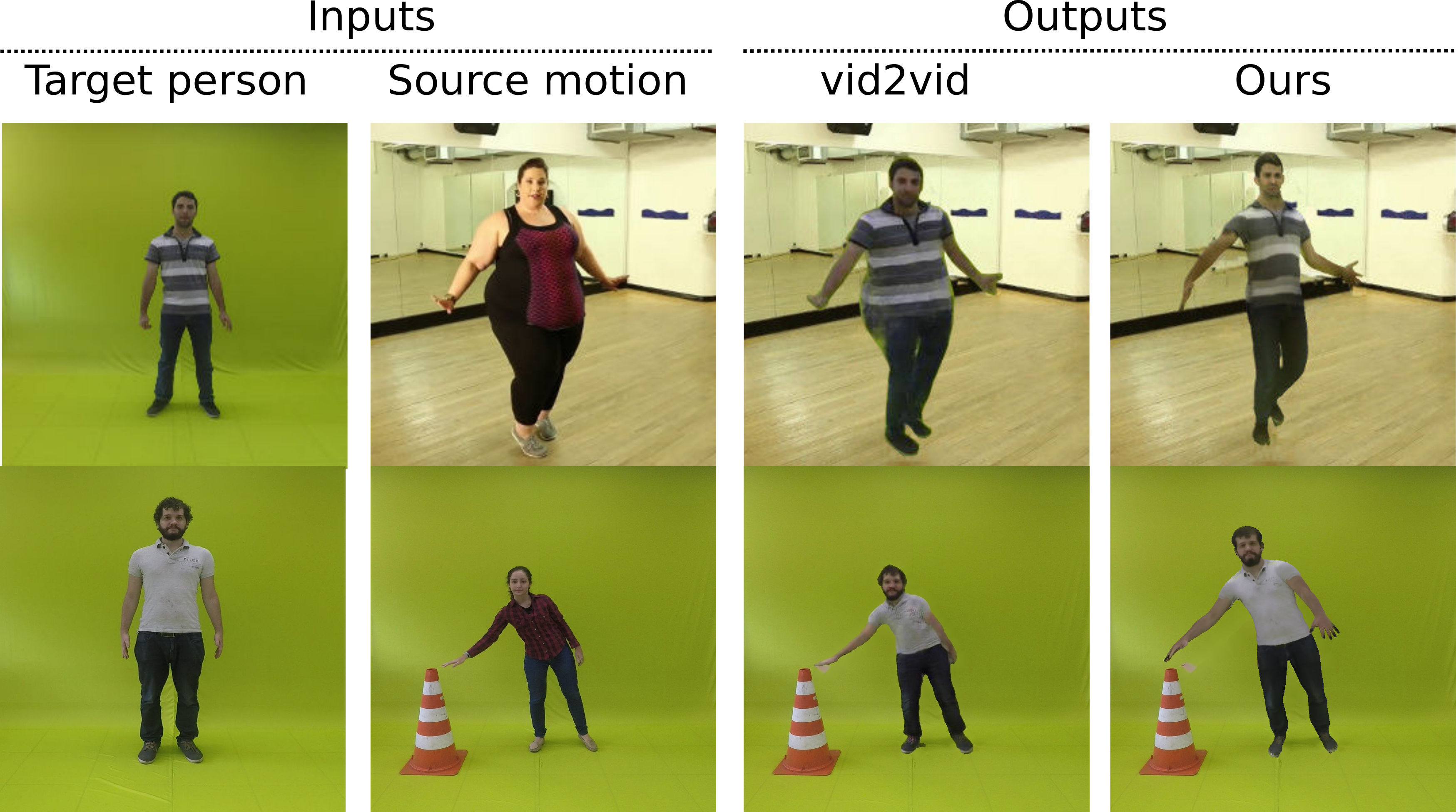}         
	\caption{\small Motion and appearance transfer using vid2vid \cite{wang2018vid2vid} and our formulation. From left to right: target person, source motion video with a human of different body shape, vid2vid, and our retargeting results. Note that vid2vid stretched, squeezed and shrinked the body forms whenever the transferring characters have different morphologies.}    
	\label{fig:ret_vid2vid}
\vspace{-0.2cm}
\end{figure}

Capturing human geometry and motion has been improved over the decades through model-based and learning techniques. Computer Vision and Computer Graphics communities have progressively adopted learning techniques to automate the modeling and animation process of articulated characters. We have witnessed a variety of approaches used to extract articulated character patterns and capture three-dimensional motion, shape, and appearance~\cite{2018-TOG-SFV,kanazawaHMR18,Esser_2018_CVPR,chan2018dance} from videos and still images from real actors. Despite remarkable advances in estimating 3D pose and shape, most of these methods only provide 3D meshes from the outer surfaces of objects, pose, and skeletons associated with those meshes. Even techniques such as the works of Chan~\etal~\cite{chan2018dance}, Esser~\etal~\cite{Esser_2018_CVPR}, and Wang~\etal~\cite{wang2018vid2vid} are limited to only transfer the appearance/style from one actor to another. In other words, these methods stretch or shrink the texture of a target actor to fit the texture in the movement instead of retargeting and fitting the motion into the actor (an alluring example is depicted in Figure~\ref{fig:ret_vid2vid}). 

In this paper, we propose a novel retargeting framework that unifies appearance transfer with retargeting motion from video to video by adapting a motion from one character in a video to another character. The proposed approach synthesizes a new video of a person in a different context where this person was initially recorded. In other words, given two input videos, we investigate how to synthesize a new video, where a target person from the first video is placed into a new context performing different motions from the second video. The proposed method is composed of four main components: motion estimation in the source video, body model reconstruction from target video, motion retargeting with spatio-temporal constraints, and finally image composition. By imposing spatial and temporal constraints on the joints of the characters, our method preserves features of the motion, such as feet touching the floor and hands touching a particular object. Also, our method employs an adversarial learning in the texture domain to improve textures extracted from frames and leverage details in the visual appearance of the target person. 

In this context, several recent learning-based methods have been proposed on synthesizing new pose from a source image (\eg,\cite{Lassner_GeneratingPeople,ZhaoWCLF17,ma2017pose,chan2018dance,Esser_2018_CVPR}).
 Unlike the methods \cite{chan2018dance,Esser_2018_CVPR,wang2018vid2vid} that are built on learning approaches to work in the image domain to transfer texture, our approach aims at adapting the movement from one actor to another taking into account the main factors for a moving actor: body shape, appearance and motion.


The main technical contributions of this paper are as follows: i) a marker-less human motion estimation technique that takes into account both body shape and camera pose consistencies along the video; ii) a generative adversarial network for improving visual details that works directly with texture maps to restore facial texture of human models; and iii) a unified methodology carefully designed to transfer motion and appearance from video to video that preserves the main features of the human movement and retains the visual appearance of the target character.

We demonstrate the effectiveness of our approach quantitatively and qualitatively using publicly available video sequences containing challenging problem conditions, as shown in Figure~\ref{fig:teaser}.

\section{Related Work}\label{sec:rel}


\paragraph*{3D human shape and pose estimation.} 
Several works have been proposed to estimate both the human skeleton and 3D body shape from images. 
Sigal~\etal~\cite{Sigal} compute shape by fitting a generative model (SCAPE~\cite{Anguelov_2005}) to the image silhouettes. Bogo~\etal~\cite{Bogo_2016} proposed the SMPLify method, which is a fully automated approach for estimating 3D body shape and pose from 2D joints in images. SMPLify uses a CNN to estimate 2D joint locations and then fits an SMPL body model~\cite{Loper_2015} to these joints. 
Lassner~\etal~\cite{Lassner_2017} take the curated results from SMPLify to train $91$ keypoint detectors. Some of these detectors correspond to the traditional body joints, and others correspond to locations on the surface of the body.
 Similarly, Kanazawa~\etal~\cite{kanazawaHMR18} used unpaired 2D keypoint annotations and 3D scans to train an end-to-end network to infer the 3D mesh parameters and the camera pose. Their method outperformed the works \cite{Bogo_2016, Lassner_2017} regarding 3D joint error and runtime. However, their bounding box cropping strategy, which frees 3D pose regression from having to localize the person in scale and image space, loses global information and temporal consistency required in the motion transfer.

\begin{figure*}[t]
	\includegraphics[width=1.0\linewidth]{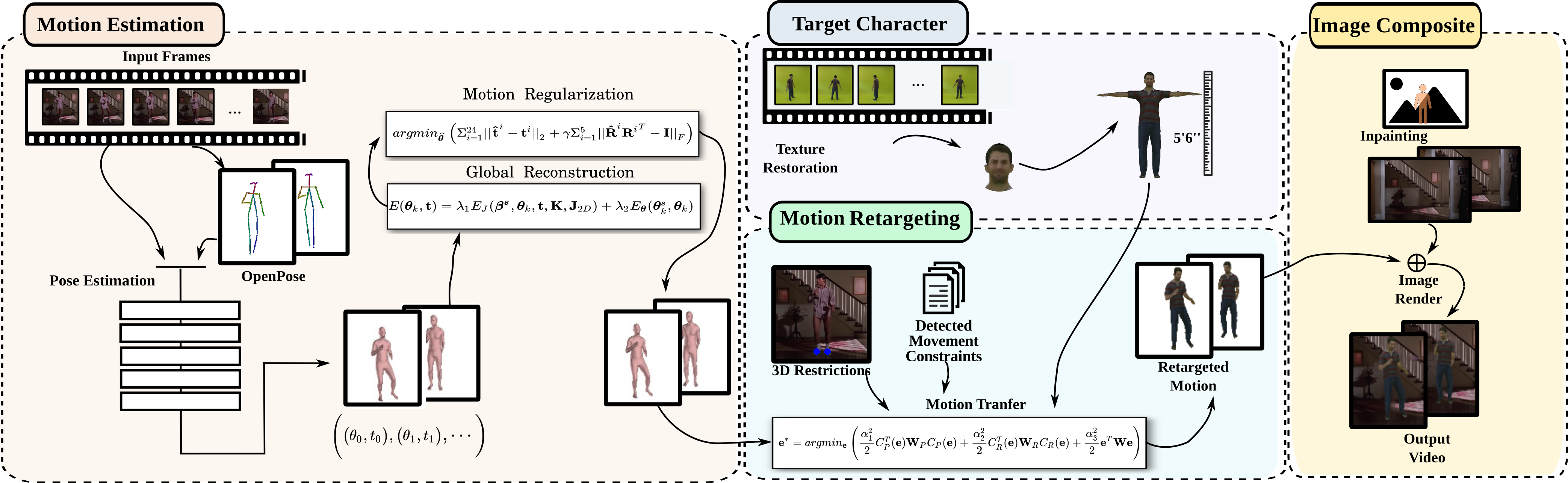}
	\caption{\small Overview of our retargeting approach that is composed of four main components: human motion estimation in the source video (first component); we retarget this motion into a different target character model (second component), considering the motion constraints (third component) and by last, we synthesize the appearance of the target character into the source video.}

	\label{fig:method}
	\vspace{-0.2cm}
\end{figure*}

\paragraph*{Retargeting motion.} Gleicher seminal work of retargeting motion~\cite{Gleicher} addressed the problem of transferring motion from one virtual actor to another with different morphologies. 
 Choi and Ko~\cite{motion_retargetting_1} pushed further Gleicher's method by presenting an online version based on inverse rate control.
 Villegas~\etal~\cite{Villegas_2018_CVPR} proposed a kinematic neural network with an adversarial cycle consistency to remove the manual step of detecting the motion constraints. In the same direction, the recent work of Peng~\etal~\cite{2018-TOG-SFV} takes a step towards automatically transferring motion between humans and virtual humanoids. Despite remarkable results in transferring different movements, these methods are limited to either virtual or textureless characters. Similarly, Aberman~\etal~\cite{retargeting_2d} proposed a 2D motion retargeting using a high-level latent motion representation. This method has the benefit of not explicitly reconstructing 3D poses and camera parameters, but it fails to transfer motions if the character walks towards the camera or with a large variation of the camera's point-of-view.

\paragraph*{Synthesizing views.} The past five years has witnessed the explosion of generative adversarial networks (GANs) to new view synthesis. GANs have emerged as promising and effective approaches to deal with the tasks of synthesizing new views, against image-based rendering approaches (\eg,\cite{image-based,image-based_2,image-based_3}). More recently, the synthesis of views is formulated as being a learning problem (\eg, \cite{CNN_for_view_synthesis_0,CNN_for_view_synthesis_1,CNN_for_view_synthesis_2,Balakrishnan,Esser_2018_CVPR}), where a distribution is estimated to sample the new views. A representative approach is the work of 
Ma~\etal~\cite{ma2017pose}, where the authors proposed to transfer the appearance of a person to a given pose in two steps. 
Similarly, Lassner~\etal~\cite{Lassner_GeneratingPeople} proposed a GAN called ClothNet. ClothNet produces people with similar pose and shape in different clothing styles given a synthetic image silhouette of a projected 3D body model. In the work of Esser~\etal~\cite{Esser_2018_CVPR}, a conditional U-Net is used to synthesize new images based on estimated edges and body joint locations. Despite the impressive results for several inputs, in most cases, these methods fail to synthesize details of the human body such as face and hands.
Recent works~\cite{Aberman_2018,chan2018dance} applied an adversarial training to map a 2D source pose to the appearance of a target subject. Although these works employ a scale-and-translate step to handle the difference in the limb proportions between the source skeleton and the target, they have still clear gaps in the motion in the test time when comparing with the motion in the training time. Wang~\etal~\cite{wang2018vid2vid} presented a general video-to-video synthesis framework based on conditional GANs to generate high-resolution and temporally consistent videos. 
 Unfortunately, these learning-based techniques transfer style and wrongly distorts characters with different morphologies (proportions or body parts' lengths). Moreover, differently from our method, these state-of-the-art approaches~\cite{Aberman_2018,chan2018dance,wang2018vid2vid} are dataset specific, \ie, they require training a different GAN for each video of the target person with different motions to perform the transferring. This training is computationally intensive and takes several days on a single GPU. Our method, for its turn, does not require a large number of images and powerful hardware for training, keeps visual details from the target character while preserving the features of the transferred motion.


\section{Retargeting Approach}

Our method can be divided into four main components. We first estimate the motion of the character in the source video. Our~\textit{motion estimation} regards essential aspects to obtain plausible character movements, such as of ensuring a common system coordinate for all image frames and temporal motion smoothness. Second, we extract the~\textit{body shape and texture} of the target character in the second video. Then, the~\textit{retargeting} component adapts the estimated movement to the body shape of the target character, while considering temporal motion consistency and the physical interactions (constraints) with the environment. Finally, the~\textit{image rendering and composition} component renders the texture (appearance), extracted from the target character, into the background of the source video. 
 Figure~\ref{fig:method} shows a schematic representation of the method pipeline. 


\subsection{Human Body and Motion Representation}

We represent the human motion by a set of translations and rotations over time of joints that specify a human skeleton. This skeleton is attached to the characters body and is defined as a hierarchy of $24$ linked joints. Each joint pose $\mbf P^i$ ($\mbf P^i \in \mathbb{SE}(3)$ is the pose of the $i$-th joint) is given by recursively rotating the joints of the skeleton tree, starting from the root joint and ending in its leaf joints (\ie, the forward kinematics denoted as FK). To represent the 3D shape of the human body, we adopted the SMPL model parametrization~\cite{Loper_2015}, which is composed of a learned human shape distribution $\mathcal{M}$, 3D joint angles ($\boldsymbol{\theta} \in \mathbb{R}^{72}$ defining 3D rotations of the skeleton joint tree), and shape coefficients $\boldsymbol{\beta} \in \mathbb{R}^{10}$ that model the proportions and dimensions of the human body.

\subsection{Human Motion Estimation}\label{method:motion_estimation}

\begin{figure}[!t]
	\centering\includegraphics[width=0.4\textwidth]{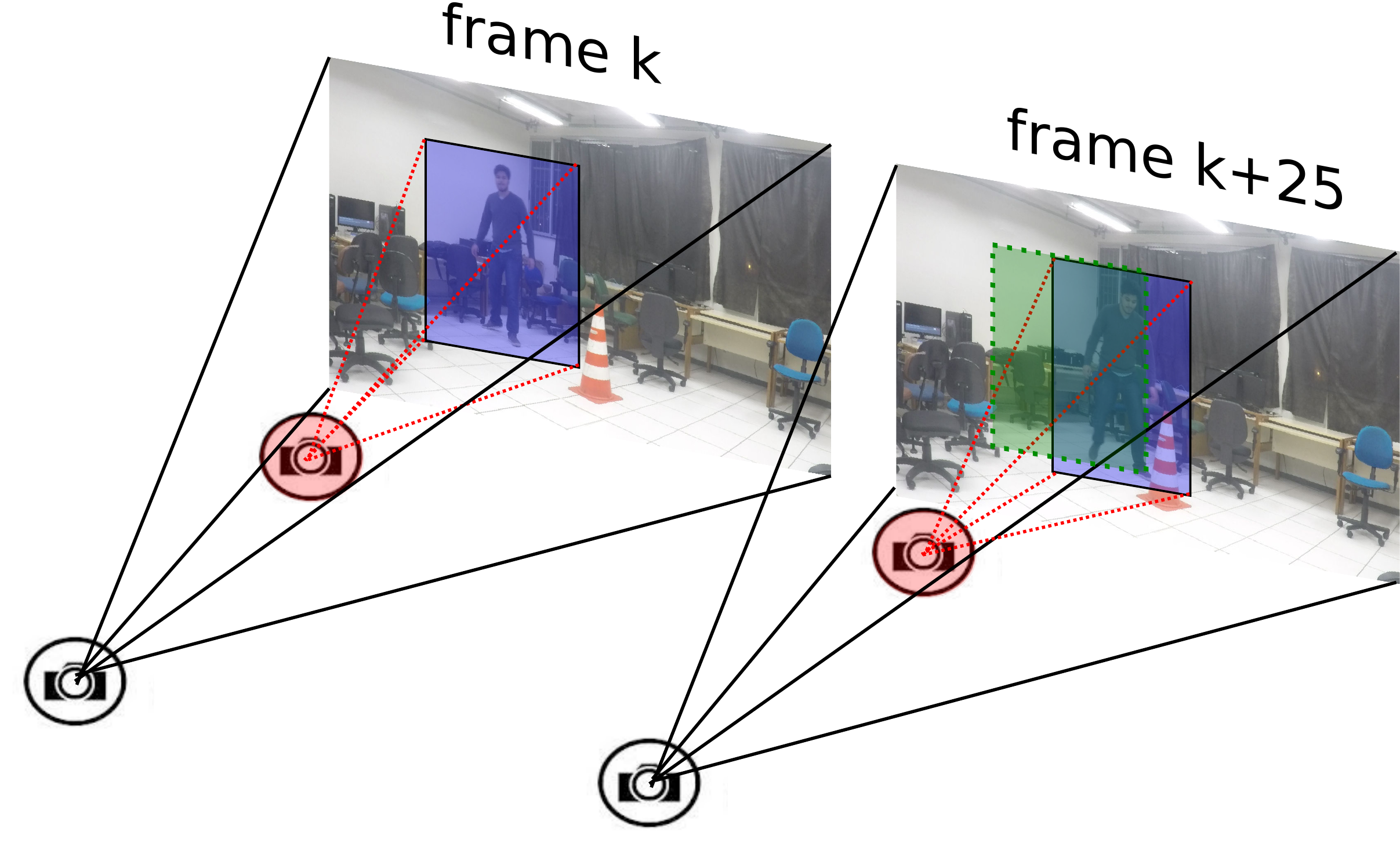}
	\caption{\small Schematic view of the motion reconstruction. Note the change of the point of view between the virtual cameras (in red) for a gap of $25$ frames (showed by the different positions of the blue and green boxes). This change is ignored by the bounding box crop, producing temporally inconsistent pose and shape estimates. }
	\label{fig:crop-camera}
	\vspace{-0.2cm}
\end{figure}

We start estimating the actor's motion in the source video. Our method builds upon the learning-based SMPL pose estimation framework of Kanazawa~\etal~\cite{kanazawaHMR18}. The human pose and shape are predicted in the coordinate system of a bounding box around the person, where a weak-perspective camera model is adopted as shown in Figure~\ref{fig:crop-camera}. 
This bounding box normalizes the person in size and position, as also noted in \cite{Mehta_2017}, which frees 3D pose estimation from the burden of computing the scale factor (between the body shape to the camera distance) and the location in the image. However, this incurs in a loss of temporal pose consistency required in the motion transfer. This also often leads to wrong body shape estimates for each frame, which should be constant along the video. 

In order to overcome these issues, we map the initial pose estimation using virtual camera coordinates, as illustrated in Figure~\ref{fig:crop-camera}. For that, our motion estimation minimizes an energy function with two terms: 
\begin{equation}\label{eq:pose}
E(\boldsymbol{\theta}_k,\mbf t) = \lambda_1E_{J}(\boldsymbol{\beta^s},\boldsymbol{\theta}_k,\mbf t,\mbf K,\mbf{J}_{2D}) + \lambda_2E_{\boldsymbol{\theta}}(\boldsymbol{\theta}^s_k,\boldsymbol{\theta}_k),
\end{equation}
\noindent where $\mbf t \in \mathbb{R}^3$ is the translation, $\mbf K \in \mathbb{R}^{3\times 3}$ is the camera intrinsic matrix, $\mbf{J}_{2D}$ is the projections of the joints in the reconstruction of \cite{kanazawaHMR18}, and $\lambda_1$, $\lambda_2$ are scaling weights. The first term encourages the pose projections of the joints to remain in the same locations into the common reference coordinate system. The second term favors maintaining the joints' angles configuration, while reinforcing the adopted character shape to have averaged shape coefficients ($\boldsymbol{\beta^s}$) of the entire video. Finally, the human model pose in each frame is then obtained with the forward kinematics (FK) in the skeleton tree:

\begin{equation}
\left(\mbf P^0_{k} ~ \mbf P^1_{k} ~ \ldots~ \mbf P^{23}_{k}\right) = \mbox{FK}(\mathcal{M},\boldsymbol{\beta^s},\boldsymbol{\theta_k}),
\end{equation}
\noindent where $\mbf P_i^{k} = [\mbox{FK}(\mathcal{M},\boldsymbol{\beta}^s,\boldsymbol{\theta}_k^s)]_i$ is the pose of the joint $i^{th}$ at frame $k$. Thus, we define the motion of each joint $i$ as the set of successive poses in the frames $\mbf M^i = [\mbf P_1^{i} ~ \mbf P_2^{i} ~ ... ~ \mbf P_n^{i}]$.

\paragraph*{Motion Regularization.} Since the character poses are estimated frame-by-frame, the resulting motion might present some shaking motion with high-frequency artifacts in some short sequences of the video. To reduce these effects, we apply a motion reconstruction to seek a new set of joint angles $\widehat{\boldsymbol{\theta}^s}$ that creates a smoother character motion. 
We compute a smoother configuration for the joints by minimizing the following cost of inverse kinematics (IK) in the joint positions and end-effectors orientations $\mbf{M}^i$:
\begin{equation}\label{eq:inversek}
\hspace{-0.05cm}
\argmin_{\widehat{\mth}}\left(\Sigma_{i=1}^{24}||\mbf{\hat{t}}^i - \mbf t^i||_2 + \gamma\Sigma_{i=1}^5||\mbf{\hat{R}}^i{\mbf{R}^i}^T - \mbf{I}||_F \right),
\end{equation}

\noindent where $\mbf{\widehat{P}}^i = [\mbf{\hat{R}}^i ~ \mbf{\hat{t}}^i]$ is given by the forward kinematics $\mbf P^i_{k} = [\mbox{FK}(\mathcal{M},\boldsymbol{\beta}^s,\boldsymbol{\theta}^s_k)]_i$ with unknown joint angles $\boldsymbol{\theta}^s_k$, $||.||_F$ is the Frobenius norm of the orientation error, and $\gamma$ the scaling factor between the position of all joints and orientation of the end-effectors (\ie, feet, hands, and head). This reconstruction strategy removes high-frequency artifacts of the motion while maintaining the main movement features of the body end-effectors.


\subsection{Target 3D Human Body Model Building}\label{method:shape_appearance}

This section presents our strategy to build the 3D model and texture of the character that is transferred to the source video (\ie, the target body model $\boldsymbol{\beta}^t$). 
 Our target reconstruction component starts with an initial 3D body model from Alldieck~\etal~\cite{alldieck2018video}. This produces a reasonable model of people in clothing from a single video in which the person is moving in an A-pose configuration. We remark that any technique, capable of creating plausible 3D human models, could be used to get this initial body model estimate in our method (\eg,~\cite{alldieck2018detailed, alldieck19cvpr,pifuSHNMKL19}). Although the good resulting 3D human model accuracy, the texture images were often blurred and lacking of details. In the following, we discuss how to mitigate the loss of detail by taking inspiration from the recent advance in generative adversarial networks.

\paragraph*{GAN Face Texture Restoration.} According to Balakrishnan~\etal~\cite{Balakrishnan}, humans are particularly good at detecting facial abnormalities such as deformations or blurring. Unfortunately, when mapping textures into a target 3D model, we lose important details mainly because of warping and interpolation artifacts.

\begin{figure}[t]
\includegraphics[width=1\linewidth]{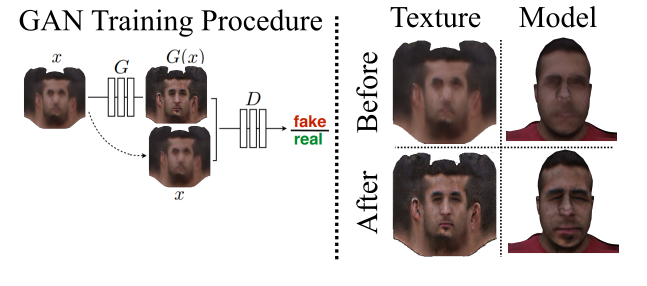}
\caption{\small Face texture enhancement strategy using a conditional GAN. An example of the restoring results in the texture domain is shown in the right column, side by side with the visualizations of the textures in the human model.}
\label{fig:gan-tex}
\vspace{-0.1cm}
\end{figure}

In order to reduce this effect, we exploit the capability of generative adversarial networks (GANs) to denoise images~\cite{pix2pix2017}. However, differently from previous works, we perform the learning directly in texture map images as shown in Figure~\ref{fig:gan-tex}. This produced better restoration results probably due to smaller geometrical variability from the texture maps compared to the appearance in the 3D body mesh. We circumvent the problem of nonexistence of a publicly available dataset of face textures, to train our GAN model, by using $500$ human face textures (real textures) from 3D.SK\footnote{https://www.3d.sk/}. We augmented the training dataset by adding noise, small rotations and blurring warped images. For the training, we adopted the conditional GAN proposed in \cite{pix2pix2017} used for image-to-image translation. Some input images of the augmented dataset and the resulting restoration can be seen in Figure \ref{fig:gan-tex} and in the supplementary material.






\subsection{Retargeting using Space-time Constraints}\label{method:retargeting_motion}

After computing the source motion ($\mbf M^i, ~\boldsymbol{\theta}^s$) and the target human 3D model ($\boldsymbol{\beta}^t$), we proceed to the motion retargeting component. The retargeting is essential to guarantee that some physical restrictions are still valid during the target character animation. In this paper, we assume that the target character has a homeomorphic skeleton structure to the source character, \ie, the main geometric differences are in terms of bone lengths or proportions. Our first goal is to retain the joint configuration of the target as close as possible of the source joint configurations, \ie, to keep the pose error $\mbf e_k = \mth_k^t - \mth_k^s$ small and then preserve the appearance of the motion whilst respecting movement constraints. A secondary objective is to keep a similar {\it movement style} in the retargeted motion over time. Thus, we propose a prediction error in 3D space to maintain the style from the original character motion:
\begin{multline}\label{eq:pred}
C_P = \sum_{k=i}^{i+n}\left(\mbox{FK}(\mathcal{M},\mbt^t,\mth_{k+1}^t) - \mbox{FK}(\mathcal{M},\mbt^t,\mth_k^t) \right.\\- 
\left.\left(\mbox{FK}(\mathcal{M},\mbt^s,\mth_{k+1}^s) - \mbox{FK}(\mathcal{M},\mbt^s,\mth_{k}^s)\right)\right).
\end{multline}

\noindent Rather than considering a full horizon cost (total number frames), we leverage only the frames belonging to a neighboring temporal window of $n$ frames equivalent to two seconds of video. This neighboring temporal window scheme allows us to track the local temporal motion style producing a motion that tends to be natural compared with a realistic looking of the estimated source motion. Only considering a local neighboring window of frames also results in a more efficient optimization. 

\begin{figure}
	\includegraphics[width=1\linewidth]{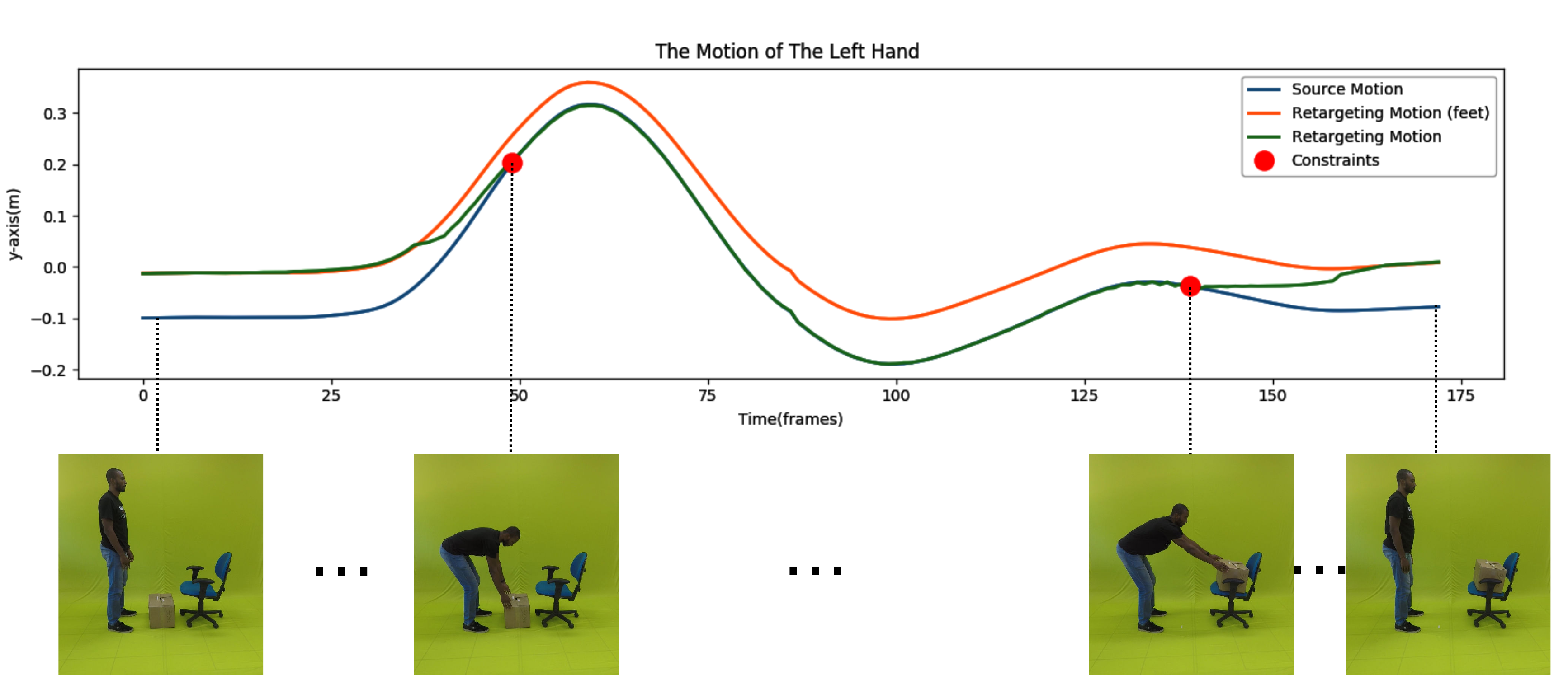}
	\caption{\small The left hand's trajectory on the y-axis when transferring the motion of {\it pick up a box} between two differently sized characters: original motion (blue line), a naive transfer without constraints at the person's hand (red line) and with constraints (green line). Frames containing motion constraints are located between the red circles.}
	\vspace{-0.1cm}
	\label{fig:grafico_retarget}
\end{figure}

\paragraph*{Spatial Motion Restrictions and Physical Interactions.} The motion constraints are used to identify key features of the original motion that must be present in the retargeted motion. The specification of these constraints typically involves only a small amount of work in comparison with the task of creating new motions. Typical constraints are, for instance, that the target character feet should be on the floor; holding hands while dancing or while grabbing/manipulating an object in the source video. Some examples of constraints are shown in Figures \ref{fig:grafico_retarget} and \ref{result:retarget-results}, where the characters are placing their left hand in a box or over a cone object.

Our method is capable of adapting to such situations in terms of position by constraining the positioning of the end-effectors to respect a set of constraints in the frame $k$ given by the joint poses $\mbf P_{R} = \left(\mbf P^j ~ \mbf P^m ~ \ldots~ \mbf P^n\right)$ as: 
\begin{equation}\label{eq:constr}
  C_{R} = \Sigma_{i}\left([\mbox{FK}(\mathcal{M},\mbt^t,\mth_{k}^t)]_i - \mbf [\mbf P_{R}]_i \right).
\end{equation}




\paragraph*{Space-time Cost Error Optimization.} The final motion retargeting cost combines the source motion {\it appearance} with the different shape and restrictions of the target character using equations (\ref{eq:pred}) and (\ref{eq:constr}):

\begin{multline}
\mbf e^* = \argmin_\mbf e \left(\frac{\alpha_1^2}{2}C_P^T(\mbf e)\mbf W_PC_P(\mbf e) + \right. \\ \left. +\frac{\alpha_2^2}{2}C_R^T(\mbf e)\mbf W_RC_R(\mbf e) + \frac{\alpha_3^2}{2}\mbf e^T\mbf W\mbf e\right),
\end{multline}

\noindent where $\mbf e = (\mbf e_{k+1}, \ldots, \mbf e_{k+n})^T$, $n$ the number of frames considered in the retargeting, $\alpha_1$, $\alpha_2$ e $\alpha_3$ are the contributions for the different error terms and, $\mbf W_P, \mbf W_R \mbox{ and } \mbf W$ are diagonal matrices of weights for the prediction, restrictions and motion similarity terms. Each of these weight matrices are set such as to penalize more the errors in joints that are closer to the root joint. We minimize this cost function with a gradient-based NLSQ iterative optimization scheme, where the Jacobians are computed using automatic differentiation for each degree of freedom. The optimization stops when either the error tolerance or the maximum number of iterations are reached. An example of the retarget motion trajectory of the left hand of our approach is shown in Figure~\ref{fig:grafico_retarget}. Note the smooth motion adaptation produced by the retargeting with the restrictions in frames $47$ and $138$ (green line) when the character's hand was touching the box. 

\subsection{Model Rendering and Image Compositing}\label{method:render}

The last step of your framework composed the rendered target character and the source background. For that, we first segment the source image into a background layer using, as a mask, the projection of our computed model with a dilation. Next, the background is filled with the method proposed by Criminisi~\etal~\cite{based-inpainting} to ensure temporal smooth to the final inpainting. We compute the final pixel color value as the median value between the neighboring frames. Finally, the background and our render of the model are composed using Poisson blender~\cite{Perez} to illumination adjustment. We remark that we tested different inpainting formulations, comprising deep learning-based methods presented in \cite{yu2018generative,wang2018videoinp}. Our experiments showed that although these deep learning-based methods synthesize plausible pixels for each frame, the adopted inpainting strategy has better results considering the visual and spatio-temporal consistency between the frames. Furthermore, \cite{wang2018videoinp} requires a static mask in the video and this is too restrictive to our problem.  

\section{Experiments and Results}

\paragraph*{Video Sequences.} The selected videos cover a variety of representative conditions to the problem, such as different types of motion, lighting conditions and background, actors morphologies and videos used by previous works. All four sequences contain different subjects and types of motion constraints that should be taken into account in order to synthesize plausible videos. A short description of these videos is as follows: i)~\textbf{alfonso-ribeiro}\footnote{https://www.youtube.com/watch?v=pbSCWgZQf\_g}: This video has strong illumination changes and fast character motions, with a $1.67$ meters height male character dancing.
 The restrictions are mostly in the dancers' feet; ii)~\textbf{joao-pedro}: Video with moderate speed motions and with a static background, where a $1.80$ meters height male character is walking and interacting with a cone in the floor. The motion constraints are in the feet and hands;
iii)~\textbf{tom-cruise}\footnote{https://www.youtube.com/watch?v=IUj79ScZJTo}: This video contains motions with moderate speed but with displacements in all directions in the scene, where a $1.70$ meters height male character is pretending to sing while dancing. 
 The motion restrictions are in the dancer's feet; iv)~\textbf{bruno-mars}\footnote{https://www.youtube.com/watch?v=PMivT7MJ41M}: Video with fast motions where a $1.65$ meters height male character is dancing, with partial occlusions of arms and feet. The restrictions are in the dancer's feet. This sequence was also used by \cite{chan2018dance}.


In order to build the target person shape and appearance, we used videos of People-Snapshot dataset~\cite{alldieck2018video} and videos of actors recorded with a Point Grey camera. These sequences were captured with the camera fixed at a distance of three meters from the characters. 


\vspace{-0.2cm}

\paragraph*{Parameters Setting and GAN Training.} 

We set $\lambda_1 = 10^{-6}$ and $\lambda_2 = 10^{-2}$ in the motion estimation. In the reconstruction and retargeting steps, we used $\gamma = 10$, $\alpha_1 = 10$, $\alpha_2 = 5$ and $\alpha_3 = 1$. Our textured dataset augmentation process was performed applying random small pixel translations (between $-15$ and $15$ pixels) and random rotations (between $-25$ and $25$ degrees) for the same image. Each original texture map was replicated twenty times with random transformations resulting in a training set of $10,000$ images.
As suggested by Isola~\etal~\cite{pix2pix2017}, our training loss is a traditional GAN loss combined with a $L1$ loss to reduce including visual artifacts. We used a factor $\lambda=500$ (conversely to $\lambda=100$ employed in \cite{pix2pix2017}) in order to avoid including visual artifacts. The other remaining training parameters were the Adam solver with learning rate of $0.0002$ and momentum $\beta_{1} = 0.5$, $\beta_{2} = 0.999$.

\paragraph*{Baseline and Metrics.}
We used the V-Unet proposed by Esser~\etal~\cite{Esser_2018_CVPR} as a baseline. The baseline choice follows two main reasons: i) Most of the related work to our approach are in the area of image-to-image translation using conditional GAN’s and the V-Unet is a recent state-of-the-art technique that represents this class of approaches; ii) Recent state-of-the-art techniques such as \cite{Aberman_2018,chan2018dance,wang2018vid2vid} are dataset specific, \ie, they need to train a GAN for each video where the target subject is performing a large set of different poses. This training has a consequent computational effort, which can last several days. Furthermore, they did not provide the code or training data making the comparison impractical. Lastly, we recall that these methods are limited to transfer solely style and suffers from the structure issue previously discussed in Section~\ref{sec:rel} and shown in Figure~\ref{fig:ret_vid2vid}.

Due to the lack of ground truth data for retargeting between two different video subjects, we adopted the same quantitative visual evaluation metrics of \cite{chan2018dance}. Thus, we measure the appearance quality using Structural Similarity (SSIM)~\cite{Wang04imagequality} and Learned Perceptual Image Patch Similarity (LPIPS)~\cite{zhang2018perceptual} between consecutive frames. For a fair comparison, the final image composition of the V-Unet uses the same background inpainting and post-process used in our method. We also report the average number of missed 2D joints' detections from OpenPose~\cite{openpose1,openpose2,openpose3} in the produced videos. Table~\ref{table:results} shows the quantitative appearance metrics and Figure~\ref{fig:retarget_results} depicts some frames for all video sequences.

\subsection{Discussion}
\begin{table}[!t]
\centering
\caption{\small Visual quantitative metrics of our method and V-Unet.}
\label{table:results}
\resizebox{\columnwidth}{!}{
\begin{tabular}{@{}crrrrrrrrc@{}}
\toprule 
Video sequence & \multicolumn{2}{c}{SSIM$^1$ } & \phantom{abc} & \multicolumn{2}{c}{LPIPS$^2$} & \phantom{abc} & \multicolumn{2}{c}{Missed detections$^3$}     & \phantom{abc} \\ \cmidrule{2-3} \cmidrule{5-6} \cmidrule{8-9}
\multicolumn{1}{r}{}                      & V-Unet       & \multicolumn{1}{r}{Ours}    && V-Unet    & \multicolumn{1}{r}{Ours}      && V-Unet    & \multicolumn{1}{r}{Ours} \\ \midrule 

\multicolumn{1}{l}{alfonso-ribeiro}    & $0.834$       & \multicolumn{1}{r}{$\mathbf{0.837}$}    && $0.137$     & \multicolumn{1}{r}{$\mathbf{0.126}$}     && $0.554$ & \multicolumn{1}{r}{$\mathbf{0.342}$} \\

\multicolumn{1}{l}{joao-pedro}    & $0.980$       & \multicolumn{1}{r}{$\mathbf{0.987}$}    && $0.018$     & \multicolumn{1}{r}{$\mathbf{0.009}$}     && $0.596$ & \multicolumn{1}{r}{$\mathbf{0.513}$} \\

\multicolumn{1}{l}{tom-cruise}    & $0.986$       & \multicolumn{1}{r}{$\mathbf{0.988}$}    && $0.013$     & \multicolumn{1}{r}{$\mathbf{0.008}$}     && $0.867$ & \multicolumn{1}{r}{$\mathbf{0.832}$} \\

\multicolumn{1}{l}{bruno-mars}    & $0.950$       & \multicolumn{1}{r}{$\mathbf{0.962}$}    && $0.044$     & \multicolumn{1}{r}{$\mathbf{0.035}$}     && $\mathbf{0.245}$ & \multicolumn{1}{r}{$0.301$} \\

\multicolumn{1}{l}{}    & \multicolumn{2}{c}{\scriptsize{$^1$\textit{Better closer to 1.}}} & & \multicolumn{2}{c}{\scriptsize{$^2$\textit{Better closer to 0.}}} & & \multicolumn{2}{c}{\scriptsize{$^3$\textit{Better closer to 0.}}} \\
\bottomrule 
\end{tabular}
}
\vspace{-0.1cm}
\end{table}

\begin{figure*}[h!]
	\centering

	\includegraphics[width=0.8\textwidth]{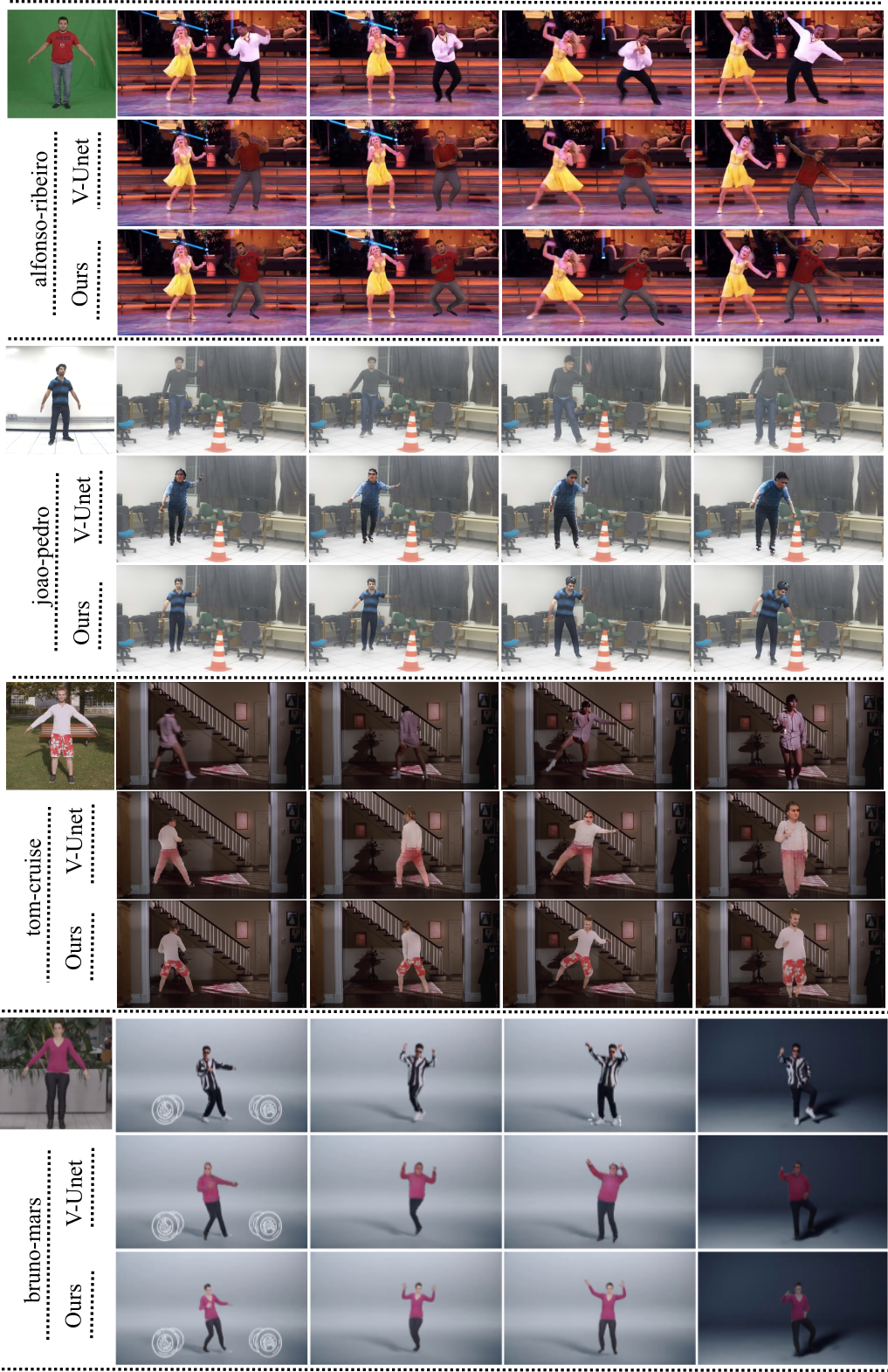} 
	\caption{\small Qualitative retargeting results using video sequences with different types of motion, lighting conditions, background and actors morphologies. In each sequence: First row: target person and motion source; Second row: V-Unet result; Third row: Our method.}	
	\label{fig:retarget_results}
\end{figure*}

\paragraph*{Visual Appearance Analysis.} It can be seen from the quantitative and qualitative results for all sequences that the proposed method leads the performance in both SSIM and LPIPS metrics. Furthermore, Figure~\ref{fig:retarget_results} shows that our method presents a much richer and detailed visual appearance of the target character than when using V-Unet. One can easily recognize the person from the target video (top left image) in the samples of the retargeting video (third rows for each video). 


To assets in which extent our texture denoising contributes to the success of our approach in retaining the visual appearance of the target, we tested our network restoration in several face textures from People-Snapshot dataset~\cite{alldieck2018video} and in our generated target human models. Figure~\ref{fig:gan-tex} shows some results after applying the denoising GAN to improve the texture for typical faces. We provide some additional denoising results in the supplementary material. 



\paragraph*{Shape and Motion analysis.} We show in Figure~\ref{fig:retarget_results} some resulting frames for all four video sequences. 
Since V-Unet uses a global pose normalization, it resizes the source image to approximate scale and location of the target person and, then, it was not able to maintain the length of the limbs during the transferring. As a result, the limbs were stretched to fit the source shape. Conversely, the proposed approach did not stretch or shrink the body forms because it regards shape, appearance as well as the motion constraints to define the form of the retarget character.

\begin{figure}
	\includegraphics[width=1\linewidth]{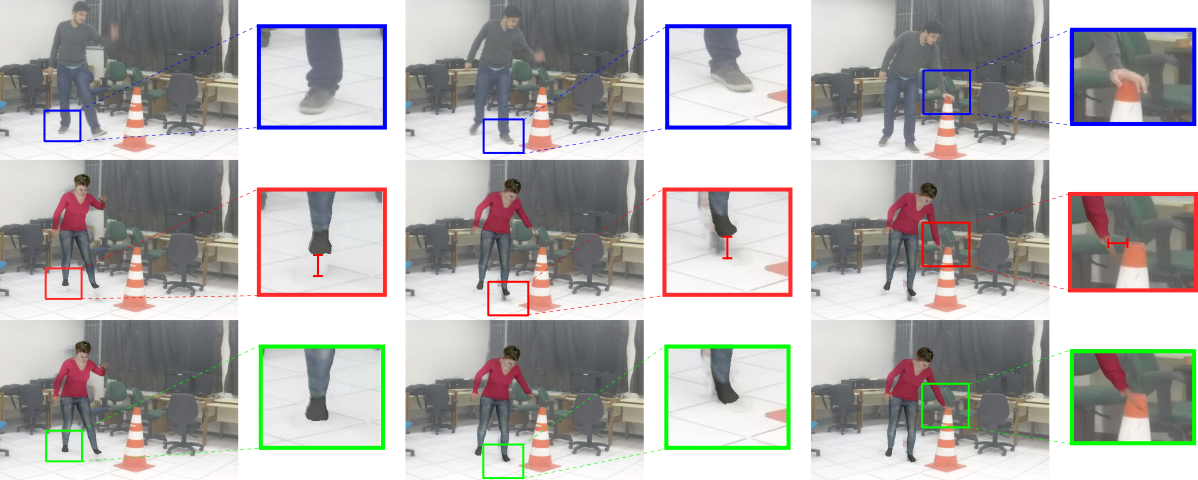}
	\caption{\small Video sequence results with motion constraints. \textbf{Top row:} source video. \textbf{Middle row}: results using a naive transferring without retargeting constraints. \textbf{Bottom row}: obtained results with our method considering the motion retargeting constraints.}
	\label{result:retarget-results}
	\vspace{-0.1cm}
\end{figure}


In terms of motion reconstruction, our method also outperformed V-Unet. For instance, V-Unet clearly failed to place the target's feet on the right position in the last frame of alfonso-ribeiro results shown in Figure~\ref{fig:retarget_results}. Due to the temporal reconstruction, our method was able to transfer the target to the correct pose. Additionally, these results reinforce the capability of our method to impose different space-time constraints to the retargeting motion. As shown on the frames of Figure~\ref{fig:retarget_results}, different motions are adapted to fit the proportions of the target person and to keep the constraints of the motion, such as of placing the hand on cone object, as illustrated in the results from the sequence joao-pedro in Figures \ref{result:retarget-results} and \ref{fig:retarget_results} for two distinct target characters. We provide additional results in the supplementary material. 

%
%
%

\section{Conclusions}

In this paper, we proposed a complete retargeting framework that incorporates different strategies to extract and to transfer human motion, shape, and appearance between two real characters in monocular videos. Differently from classic retargeting methods that use either appearance or motion information, the proposed framework takes into account simultaneously four important factors to retargeting, \ie, pose, shape, appearance, and features of the motion.

We performed real transferring experiments on publicly available videos. Our approach outperforms V-Unet in terms of both appearance metrics (SSIM and LPIPS) and number of missed joints' detections when estimating the skeleton. Our results suggest that retarget strategies based on image-to-image translation are not powerful enough to retarget motions while keeping the desired constraints of the motion and shape/appearance. Future work directions include automatically detecting the retargeting motion constraints in the videos, as well as improving the appearance restoration and transferring beyond the face texture maps. Another interesting topic would be to improve the scene compositing (\eg, estimating the scene illumination) for a more realistic rendering.


\paragraph*{Acknowledgments.} The authors would like to thank CAPES, CNPq, FAPEMIG, and ATMOSPHERE PROJECT for funding different parts of this work. We also thank NVIDIA Corporation for the donation of a Titan XP GPU used in this research.

\clearpage

{\small
	\bibliographystyle{ieee}
	\bibliography{wacv_gomes2020}
}
\end{document}